\documentclass[letterpaper, 10 pt, journal, twoside]{IEEEtran} 

\usepackage{amsmath,amsfonts}
\usepackage{algorithmic}
\usepackage{algorithm}
\usepackage{array}
\usepackage[caption=false,font=normalsize,labelfont=sf,textfont=sf]{subfig}

\usepackage{textcomp}
\usepackage{stfloats}
\usepackage{url}
\usepackage{verbatim}
\usepackage{graphicx}
\usepackage{makecell}
\usepackage{xcolor}
\usepackage{booktabs,multirow,pifont}
\usepackage{cite}
\usepackage{float}
\usepackage{siunitx}
\usepackage{colortbl}
\usepackage{rotating}     
\usepackage{multirow}     

\usepackage[font=footnotesize]{caption}
\usepackage{titlesec}
\usepackage{subcaption}

\begin{document}

\title{EmbodiedDiffusion: End-to-End Traversability-Guided \\ Visual Diffusion for Heterogeneous Robot Navigation}

\author{%
    Iana~Zhura$^{1}$,
    Sausar~Karaf$^{1}$,
    Faryal~Batool$^{1}$,
    Nipun~Dhananjaya~Weerakkodi~Mudalige$^{1}$,
    Valerii~Serpiva$^{1}$,\\
    Ali~Alridha~Abdulkarim$^{1}$,
    Aleksey~Fedoseev$^{1}$,
    Didar~Seyidov$^{1}$,
    Hajira~Amjad$^{1}$, and
    Dzmitry~Tsetserukou$^{1}$%
    \thanks{Manuscript received: February 13, 2026; Revised: May 27, 2026; Accepted: July 21, 2026.}%
    \thanks{This paper was recommended for publication by Editor Olivier Stasse upon evaluation of the Associate Editor and Reviewers' comments.
    Research reported in this publication was financially supported by the RSF grant No. 24-41-02039.}%
    \thanks{$^{1}$The authors are with the Intelligent Space Robotics Laboratory, Center for Digital Engineering, Skolkovo Institute of Science and Technology, Moscow, Russia.
    {\tt\footnotesize \{iana.zhura, sausar.karaf, faryal.batool, nipun.weerakkodi, valerii.serpiva, ali.abdulkarim, aleksey.fedoseev, didar.seyidov, hajira.amjad, d.tsetserukou\}@skoltech.ru}}%
    \thanks{Digital Object Identifier (DOI): see top of this page.}%
}

\markboth{IEEE ROBOTICS AND AUTOMATION LETTERS. PREPRINT VERSION. ACCEPTED AUGUST, 2026}%
{Zhura \MakeLowercase{\textit{et al.}}: EmbodiedDiffusion: End-to-End Traversability-Guided Visual
Diffusion for Heterogeneous Robot Navigation}

\maketitle

\begin{abstract}
Visual traversability estimation is central to autonomous 
navigation, yet most approaches either rely on prompt-driven Vision-Language Models (VLMs) or decouple traversability 
from trajectory planning, requiring separate planners with 
heavy mapping, manual tuning, and extended deployment 
time. We propose \emph{EmbodiedDiffusion}, a diffusion-based framework that simultaneously predicts traversability 
maps and generates feasible trajectories from RGB images 
using planner-free synthetic supervision and embodiment 
conditioning for cross-platform transfer. The framework 
distills category-level traversability semantics from a 
VLM teacher into a lightweight student model
during training, enabling prompt-free, real-time inference 
at deployment. A modular FiLM-based conditioning mechanism 
isolates embodiment-specific reasoning into a compact 
trainable subset of the network, allowing rapid adaptation 
to new robot platforms without retraining the visual 
backbone or the trajectory diffusion model. Across indoor 
environments with quadruped and aerial robots, 
\emph{EmbodiedDiffusion} achieves 80--100\% navigation success in the full-data regime with
real-time inference 90\,ms and adapts to new platforms using
only $\sim$10\,min of visual data collection, demonstrating scalable, unified traversability 
reasoning and trajectory generation for heterogeneous 
robots.
\end{abstract}


\IEEEpeerreviewmaketitle
\section{Introduction}

\IEEEPARstart{I}{ndoor} navigation enables warehouse automation, inspection, 
and search-and-rescue~\cite{gomez23,weerakoon2023adventr}, 
requiring simultaneous traversability reasoning and 
trajectory generation in cluttered environments where small 
errors cause failure.

\begin{figure}[t]
\centering
\includegraphics[width=\linewidth]{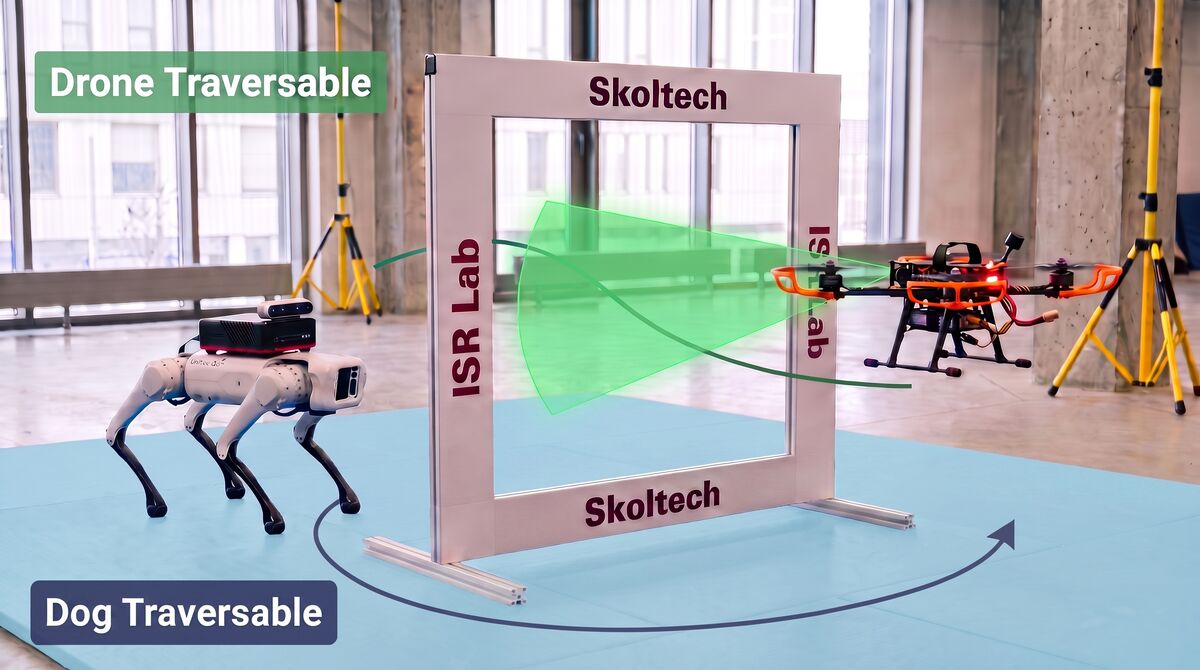}
\caption{Cross-embodiment traversability and trajectory generation from a 
shared visual backbone. A single network conditioned on the robot state 
vector via FiLM produces embodiment-specific behavior from the same RGB 
scene: the quadruped (blue) routes around the gate while the aerial robot 
(green) flies through it, with only a lightweight trainable subset adapted 
per platform.}
\label{fig:teaser_cross_embodiment}
\end{figure}

The cross-embodiment setup is illustrated in Fig.~\ref{fig:teaser_cross_embodiment}. Classical geometric pipelines rely on explicit metric maps 
and careful tuning, making them brittle under occlusions 
and irregular geometry. Self-supervised visual traversability 
methods~\cite{frey2023fast,gasparino2024wayfaster} reduce 
map dependence but remain tightly coupled to the sensing 
modality and embodiment that generated the supervision, 
requiring retraining for each new platform. 
VLMs offer greater flexibility 
but require per-scene, per-embodiment prompt tuning at 
deployment, a significant overhead in practice~\cite{sahu2025anytraverse, vla}. Critically, most existing 
learned navigation approaches either memorize expert 
trajectory datasets or depend on planner-generated ground 
truth, limiting generalization to distributions not seen 
during training~\cite{von-platen-etal-2022-diffusers, ldp}.

We propose \textit{EmbodiedDiffusion}, a diffusion-based 
framework that simultaneously predicts traversability maps 
and generates feasible trajectories from a single RGB 
image, trained without expert demonstrations using 
planner-free synthetic supervision. VLM knowledge is 
distilled offline into a compact student model that performs
prompt-free inference, eliminating per-scene tuning 
while preserving the semantic richness of the VLM. A modular FiLM-based embodiment 
conditioning mechanism isolates platform-specific 
reasoning into a lightweight trainable subset, keeping 
the visual backbone and trajectory diffusion model frozen 
to enable rapid cross-platform adaptation from only 
10\,min of data.


Our main contributions can thus be summarized as follows:

\begin{itemize}

\item \textbf{Unified traversability and trajectory 
generation from RGB images.} A single diffusion model that 
simultaneously predicts traversability and generates 
feasible trajectories from RGB images, without expert 
demonstrations or mapping. VLM supervision is distilled 
offline into a compact prompt-free student model running at 
90\,ms on edge hardware.

\item \textbf{Planner-free synthetic supervision with 
traversability regularization.} Trajectories are learned 
from geometrically generated heatmaps, where $\mathcal{L}_{\text{trav}}$ concentrates the probability mass in visually safe regions. Learning the trajectory distribution rather than memorizing specific paths yields 
stronger generalization and outperforms expert-supervised 
baselines in obstacle clearance.

\item \textbf{Modular embodiment-conditioned adaptation.} 
A FiLM-based mechanism that freezes the DINOv2 backbone 
and diffusion UNet while retraining only the lightweight 
ViT adapter, state encoder, and decoder, enabling new 
platform adaptation from 10\,min of visual data.

\item \textbf{Heterogeneous real-world and simulation-based validation.} 
We provide a complete ROS2 navigation framework validated on a quadruped 
(Unitree Go1) and a custom aerial platform, alongside a Unity simulation 
environment developed for safe UAV testing, achieving 80--100\% success 
in the full-data regime at 90\,ms inference. Code, dataset, and simulation 
environments are available at \url{https://iana-zhura.github.io/EmbodiedDiffusion/}.

\end{itemize}

\section{Related Work}

\subsection{Traversability Estimation from Vision}
Self-supervised traversability estimation derives supervision from a robot's own interactions and regresses it from visual perception, using RGB~\cite{v-strong,julian2020badgr}, depth, or LiDAR inputs~\cite{chen2024alter,yoon24,lidartrav}. 
While some recent work has explored fast adaptation using pre-trained visual features for traversability prediction~\cite{frey2023fast}, these methods focus solely on traversability without trajectory generation.
More broadly, self-supervised approaches require extensive physical interaction and are tightly coupled to the embodiment and sensing configuration that generated the supervision, limiting transfer to new platforms and environments~\cite{11128264, mudalige2022dogtouch}.

VLM approaches~\cite{sahu2025anytraverse, 10960724} reduce 
this burden but require per-deployment prompt tuning for each embodiment and 
scene, which is a prohibitive cost in practice. SALON~\cite{sivaprakasam2024salon} 
automates refinement through online exploration but still incurs deployment-time overhead.

Existing methods either require physical interaction 
and deployment-time prompt engineering, or decouple 
traversability from planning, incurring mapping and 
tuning overhead. We address this by distilling 
embodiment-specific traversability cues from a VLM 
offline into a lightweight student, enabling prompt-free, unified traversability-trajectory generation at deployment.
\subsection{Diffusion-Based Motion Planning}
Diffusion models have recently been applied to motion generation, spanning quadruped pose diffusion (DiPPeST)~\cite{Stamatopoulou2024}, goal-conditioned exploration (NoMaD)~\cite{Sridhar2024}, cost-aware trajectory generation (MPD)~\cite{Carvalho2025}, image-based navigation (VENTURA~\cite{Zhang2025VENTURA}, NaviDiffusor~\cite{Zeng2025}), and UAV swarm path planning (ImpedanceDiffusion)~\cite{batool2026impedancediffusion}. Yet these methods depend on expert demonstrations or planner-generated ground truth, assume precomputed costmaps or external traversability estimates, and treat traversability and trajectory generation as disjoint stages. FlightDiffusion~\cite{serpiva2026flightdiffusion} recovers trajectories from diffusion-generated FPV video via SLAM. This dependence on generative video synthesis and downstream SLAM reconstruction incurs substantial computational cost and retains a reliance on prompting. This computational burden is not unique to learning-based approaches. Classical aerial planners encounter it as well. Artificial Potential Field (APF) methods are difficult to tune on dense 3-D point clouds. Planners that avoid such dense representations often rely instead on GPS, which is not robust in GPS-denied environments~\cite{Wang2025}. 

Critically, no diffusion-based planner unifies navigation across ground and aerial embodiments. We instead learn traversability and trajectory generation jointly from vision under planner-free synthetic supervision, without curated datasets.

\section{Technical Approach}

\subsection{Network Architecture}
\begin{figure*}[t]
    \centering
    \includegraphics[width=2\columnwidth]{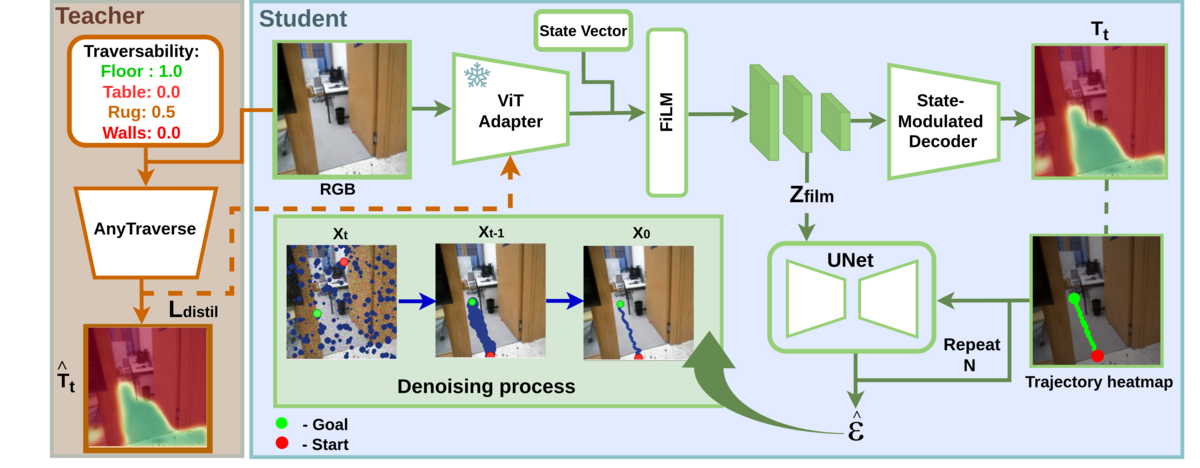}
    \caption{Two-component framework. \textit{Teacher model} (left): 
AnyTraverse~\cite{sahu2025anytraverse} produces a dense 
traversability map $\hat{T}_t$ from the RGB image and 
embodiment-specific prompts (e.g., Floor:~1.0, Walls:~0.0), 
supervising the student model via $\mathcal{L}_{\text{distil}}$. 
\textit{Student model} (right): a frozen visual encoder is 
modulated by the robot state vector through FiLM 
conditioning~\cite{perez2018film}, producing embodiment tokens $Z_{\text{film}}$ and a predicted traversability map $\hat{T}_t$ via the State-Modulated
Decoder.
$\mathbf{Z}_{\text{film}}$ additionally conditions the 
diffusion UNet via cross-attention. The UNet denoises a 
trajectory sample $x_t \to x_0$ over $N{=}20$ steps, 
conditioned on trajectory heatmap overlay and the predicted traversability map $\hat{T}_t$. The Teacher model runs 
offline during training only; at inference the student model
runs prompt-free at 90\,ms.}
    \label{fig:method}
\end{figure*}

Fig.~\ref{fig:method} illustrates the two-component framework. Our framework couples two components: (1) a traversability student model that distills embodiment-specific scene 
understanding from a VLM teacher model, and (2) a diffusion-based 
trajectory generator conditioned on the predicted 
traversability, the robot state vector, and supervised 
by synthetic trajectories.
\subsubsection{Synthetic Trajectory Supervision:}
\label{sec:synthes}
To supervise trajectory diffusion without expert 
demonstrations, we generate synthetic trajectory heatmaps 
using a simple geometric procedure. For each training 
image, we sample a start point $s$ near the image bottom 
and a goal point $g$ in a high-traversability region 
identified by the VLM teacher model. We generate smooth paths 
by sampling $K{=}5$ intermediate waypoints along 
$s{\to}g$, applying random perpendicular perturbations sampled from $\mathcal{N}(0, \sigma^2)$, smoothing via cubic splines, 
and rasterizing as thin traces on a $64{\times}64$ grid.

These heatmaps provide soft supervision that encodes
connectivity and directionality without requiring geometric 
optimality. This is deliberate: because synthetic paths 
are heuristically generated rather than expert-optimal, 
the model learns the underlying distribution of feasible 
trajectories rather than memorizing specific paths. 
Geometric path quality is further encouraged through 
path length and smoothness regularization, and start--goal 
consistency is enforced via an explicit anchoring loss, 
both described in Section~\ref{sec:objectives}.

\subsubsection{Traversability Teacher and Student Models:}
Traversability supervision comes from the AnyTraverse VLM~\cite{sahu2025anytraverse}, which provides platform-specific traversability maps as training targets and trajectory conditioning signals. 

Distillation is a deliberate design choice: the VLM teacher model is too large and prompt-dependent for real-time onboard inference. By extracting its traversability knowledge offline once per embodiment type, we obtain a compact student model that runs at 90 ms on edge hardware without any prompting at deployment.

The traversability student model uses a frozen DINOv2 ViT~\cite{oquab2024dinov2} with a lightweight adapter~\cite{vitDeblina}. 
Robot state $s_t = [x, y, z, \text{roll}, \text{pitch}, \text{yaw}] \in \mathbb{R}^6$ encodes camera pose, enabling embodiment-specific predictions via FiLM conditioning on visual features. This captures platform-specific viewpoints and reachable regions without modeling full dynamics.

The robot state vector $s_t \in \mathbb{R}^6$ is encoded 
by an MLP to produce FiLM (Feature-wise Linear 
Modulation~\cite{perez2018film}) parameters 
$(\gamma_t, \beta_t) \in \mathbb{R}^C$, which modulate 
the visual feature map 
$\mathbf{F} \in \mathbb{R}^{C \times H \times W}$ via a 
channel-wise affine transformation:
\begin{equation}
\tilde{\mathbf{F}} = \mathbf{F} \odot (1 + \gamma_t) + \beta_t ,
\end{equation}
where $\odot$ is the element-wise (Hadamard) product, 
$C$ is the number of feature channels, and $H \times W$ 
is the spatial resolution. This conditioning allows the 
same visual scene to yield embodiment-consistent 
traversability predictions.
A compact convolutional decoder produces a dense $64{\times}64$ traversability map $\hat{T}_t$.

\subsubsection{Diffusion-Based Trajectory Policy:}
Trajectory generation is performed using a conditional UNet diffusion model.
At denoising step $t$, the network predicts the injected noise $\hat{\varepsilon}_t$ conditioned on the noisy trajectory sample $x_t$, the start--goal map $S_g$, the predicted traversability $\hat{T}_t$, and embodiment tokens $Z_{\mathrm{film}}$ obtained from the same FiLM-conditioned state encoding used in the traversability model:
\begin{equation}
\hat{\varepsilon}_t = f_\theta([x_t, S_g, \hat{T}_t], t, Z_{\mathrm{film}}).
\end{equation}

Spatial conditioning ensures that denoising remains aligned with traversable regions, while token-level conditioning injects embodiment priors through cross-attention.

Trajectories are generated at inference time by iteratively denoising from $x_T \sim \mathcal{N}(0,I)$ using a cosine noise schedule.
The final output $x_0$ is a clean trajectory heatmap that is simultaneously consistent with scene geometry, traversability, and robot embodiment.

\subsection{Learning Objectives}
\label{sec:objectives}

The training objective is designed to encourage four 
complementary properties:
(i)~globally coherent trajectory structure via 
diffusion-based denoising,
(ii)~alignment of predicted trajectories with visually 
traversable regions,
(iii)~embodiment-consistent interpretation of traversability 
through distillation, and
(iv)~geometric path quality via length and smoothness 
regularization.

Let the predicted trajectory heatmap be 
$x_0 \in [0,1]^{H \times W}$, the noisy diffusion state at 
timestep $t$ be $x_t$, and the predicted traversability 
map be $\hat{T}$.

\subsubsection{Diffusion Objective:}
Trajectory generation is trained using the standard DDPM 
noise prediction objective. Given injected noise 
$\varepsilon \sim \mathcal{N}(0,I)$, the model predicts 
$\hat{\varepsilon}_t$ and minimizes:
\begin{equation}
\mathcal{L}_{\mathrm{noise}}
=
\| \hat{\varepsilon}_t - \varepsilon \|_2^2 .
\end{equation}
This objective enables the model to learn a multimodal 
distribution over feasible trajectory structures conditioned 
on visual and embodiment cues.

\subsubsection{Traversability Alignment:}
To bias trajectories toward safe regions, we define a 
spatial probability distribution over the predicted 
trajectory heatmap using a global softmax:
\begin{equation}
    p(u, v) = \frac{\exp(x_0(u, v)/\tau)}{\sum_{i,j}\exp(x_0(i, j)/\tau)},
\end{equation} where $(u, v)$ denote pixel coordinates and $\tau > 0$ is the temperature~parameter. 

The expected traversability under this distribution is:
\begin{equation}
\mathbb{E}_{p}[\hat{T}]
=
\sum_{u,v} p(u,v)\,\hat{T}(u,v),
\end{equation}
and we minimize:
\begin{equation}
\mathcal{L}_{\mathrm{trav}} = 1 - \mathbb{E}_{p}[\hat{T}],
\end{equation}
which encourages probability mass to concentrate in 
visually traversable regions without enforcing hard 
geometric constraints.

\subsubsection{Traversability Distillation:}
The traversability student model is trained via $\ell_1$ 
distillation from the teacher model prediction $T$:
\begin{equation}
\mathcal{L}_{\mathrm{distil}}
=
\frac{1}{HW}
\sum_{u,v}
\left| \hat{T}(u,v) - T(u,v) \right|,
\end{equation} where $H \times W$ denotes the spatial resolution of the heatmap ($64 \times 64$ in all experiments).

\subsubsection{Geometric Path Regularization:}

To improve geometric consistency, we apply lightweight trajectory
regularization during training. Let $\{p_i\}_{i=1}^{K}$ denote the
ordered waypoints extracted from the predicted trajectory heatmaps via
differentiable weighted centroid estimation. The geometric
regularization loss is defined as:
\begin{equation}
\mathcal{L}_g =
\sum_{i=1}^{K-1} \|p_{i+1} - p_i\|_2
\;+\;
\alpha
\sum_{i=2}^{K-1}
\|p_{i+1} - 2p_i + p_{i-1}\|_2^2,
\end{equation} where the first term penalizes trajectory path length and the second
encourages smoothness via the discrete second-order derivative.
The weighting coefficient $\alpha$ controls the smoothness strength,
and is empirically set to $\alpha = 0.1$ in all experiments based on
validation performance.

\begin{figure*}[h]
    \centering
    \includegraphics[width=1.5\columnwidth]{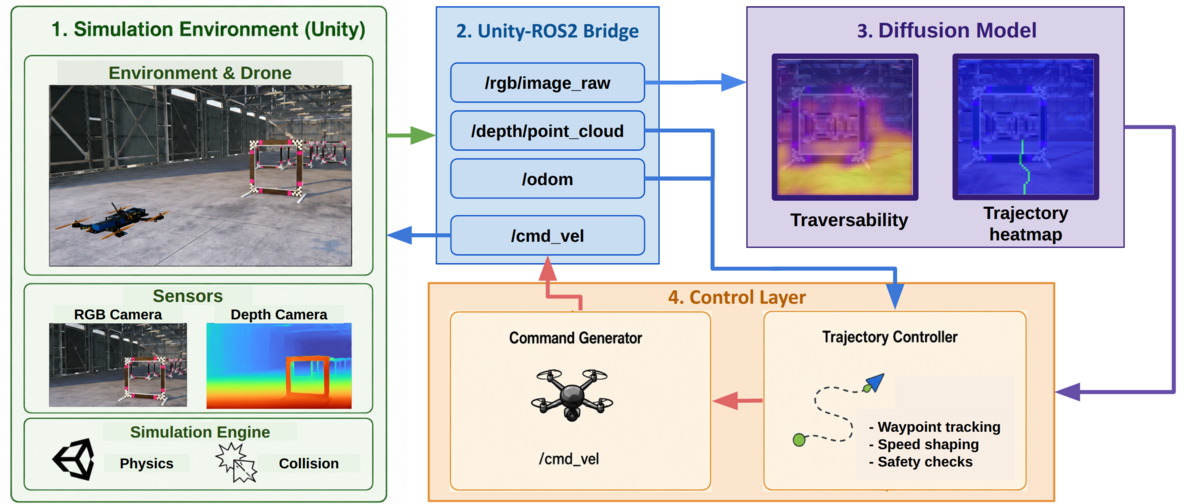}
    \caption{UAV simulation architecture. A closed-loop pipeline of four stages:
(1)~Unity renders the scene and drone with physics and collision, providing
RGB and depth streams; (2)~a Unity--ROS2 bridge exchanges sensor data and
velocity commands; (3)~the diffusion model predicts a traversability map and
trajectory heatmap from RGB; and (4)~the control layer extracts 3D waypoints
from the heatmap and depth, tracking them before issuing velocity commands
back to the simulator. Decoupling rendering from dynamics enables safe UAV
testing, with the same extraction used in real-world deployment.}
    \label{fig:uav_simulation_overview}
\end{figure*}
\subsubsection{Final Objective:}
The full training objective is given by:
\begin{equation}
\mathcal{L}
=
\mathcal{L}_{\mathrm{noise}}
+
\lambda_{\mathrm{trav}} \mathcal{L}_{\mathrm{trav}}
+
\lambda_{\mathrm{distil}} \mathcal{L}_{\mathrm{distil}}
+
\lambda_{\mathrm{g}} \mathcal{L}_{\mathrm{g}} ,
\end{equation}
where $\lambda_{\mathrm{trav}}$, $\lambda_{\mathrm{distil}}$, 
and $\lambda_{\mathrm{g}}$ are the fixed scalar weights used 
in all experiments.

\subsection{Scene Generalization}
\label{sec:general}

Traversability prompts are formulated at the category level 
rather than the scene level: broad surface categories for 
ground robots (\textit{floor}, \textit{walls}, and
\textit{rugs}) and embodiment-specific affordances for 
aerial robots (\textit{narrow gaps}, \textit{low 
ceilings}). Encoding traversability categories rather 
than specific visual instances allows the teacher model to 
transfer naturally to unseen scenes sharing the same 
semantic structure, as confirmed by the cross-scene 
experiments in Section~\ref{sec:cross-scene}.

\subsection{Modular Embodiment Adaptation}
\label{sec:embody_adapt}

The DINOv2 backbone and diffusion UNet are frozen across 
embodiments; only the ViT adapter, FiLM modulation, 
state encoder, and decoder are retrained per platform, 
enabling efficient adaptation from as few as 4{,}000 RGB 
images in 10\,min. Crucially, training data from 
multiple embodiments can be mixed within a single 
training run: the FiLM mechanism conditions visual 
features on the robot state vector, making each 
embodiment's predictions distinguishable from one another without 
requiring separate models or isolated training sets. 
This coupling between body-specific state and visual 
features ensures that the shared backbone learns 
platform-agnostic scene representations while the 
lightweight trainable subset captures the physical 
constraints of each robot, as reported in 
Section~\ref{sec:embody}.

\section{Experiments and Results}
\subsection{UAV Simulation}

For safety, we developed a Unity-based UAV simulation (Fig.~\ref{fig:uav_simulation_overview}) following Flightmare \cite{song2020flightmare}, decoupling visual rendering from dynamics. The ROS-based controller sends state to Unity, which provides RGB, depth, and point cloud data. Our diffusion pipeline generates a path mask from RGB, projected onto the point cloud to extract 3D waypoints. The same extraction method is used in real-world deployment.

\subsection{Dataset Generation}
We collected 4{,}000 RGB images per embodiment using an Intel RealSense D435i on a Unitree Go1 and a custom UAV. Robot state vectors encoding camera pose are extracted from visual odometry. For each frame, synthetic trajectories are generated automatically: starts sampled near the image bottom, goals in distant high-traversability regions. Training on 8{,}000 samples on an NVIDIA RTX~4090 requires 1.5 hours and 6\,GB of memory.

\subsection{Metrics}

We evaluate feasibility and safety against the DTG diffusion baseline~\cite{DTG}, as our contribution is visual-to-motion inference rather than trajectory optimization. Metrics include success rate, clearance (average obstacle distance), traversability cost (normalized teacher-derived cost), and Safe$_{\min}$, Safe$_{\max}$
(fraction of points with $>0.2$\,m clearance).

\subsection{Ablation Studies}

All ablation studies are performed on a held-out validation split with unseen environments, and the best configuration is subsequently used for all real-robot and simulation deployments, ensuring no test-set leakage.
\begin{figure}[t]
\centering
\includegraphics[width=\columnwidth]{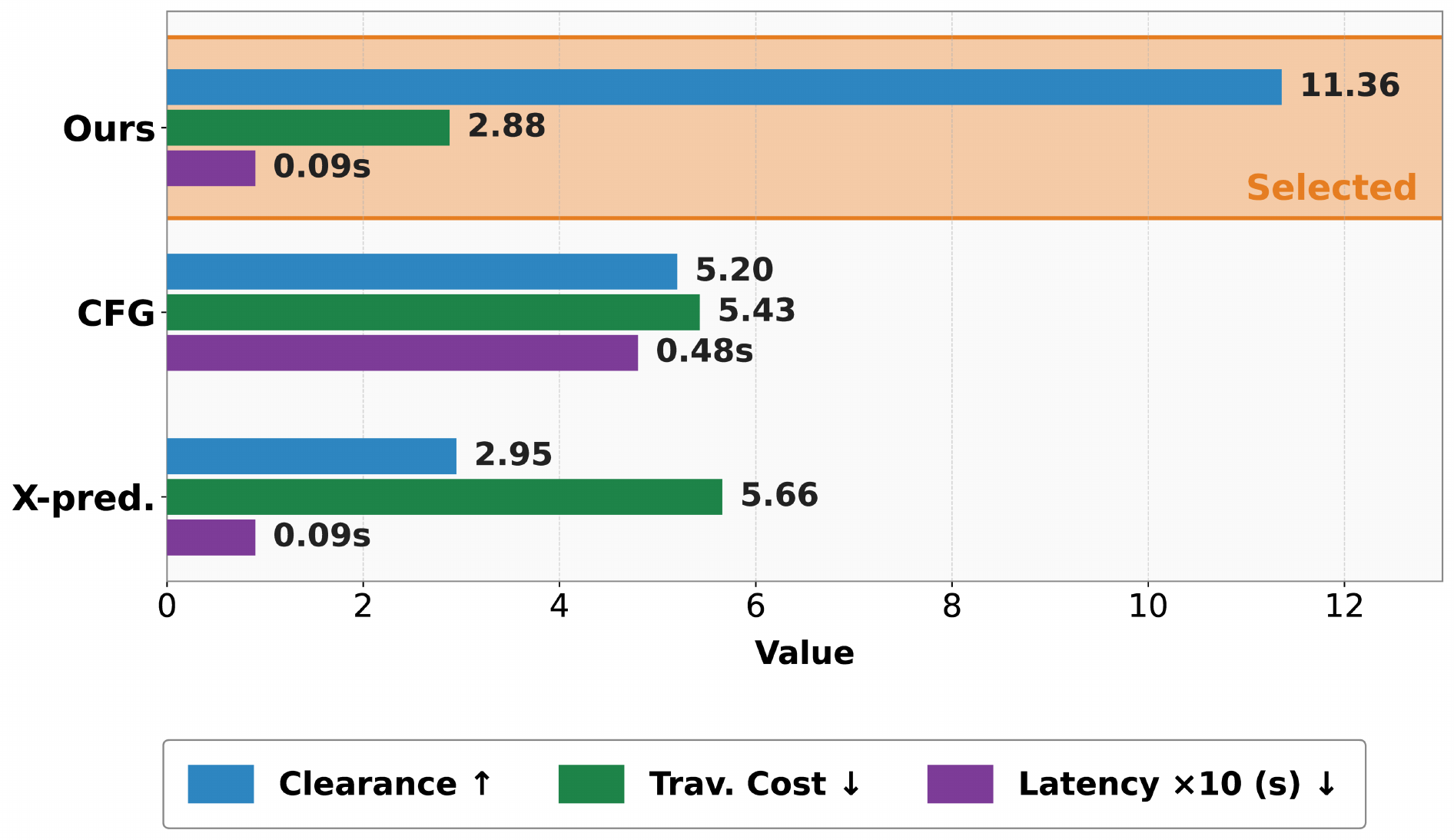}
\caption{Ablation of diffusion design choices. \textit{Ours} 
combines conditional traversability guidance with 
$\varepsilon$-prediction (noise prediction). \textit{CFG} degrades 
clearance by 54\% and slows inference $5\times$; \textit{X-pred.} 
(sample prediction, in which the clean sample $x_0$ is predicted 
directly) degrades clearance by 74\% due to overfitting on 
imperfect synthetic supervision.}
\label{fig:design_choices}
\end{figure}

\subsubsection{Influence of Different Components:}

As shown in Table~\ref{tab:path_ablation}, all components contribute positively to trajectory quality. Removing traversability input reduces clearance by 26\% and increases cost by 19\%, confirming its role in safe navigation.  Most critically, disabling conditional embedding causes the largest performance drop (clearance falls 49\%, from 11.36 to 5.83), demonstrating that embodiment-aware conditioning is essential for generating platform-specific feasible trajectories.

\begin{table}[h]
\centering
\caption{Ablation: Network Components.}
\setlength{\tabcolsep}{9pt}
\renewcommand{\arraystretch}{1.3}
\begin{tabular}{lcccc}
\toprule
\textbf{Variant} & \textbf{Clear.\,$\uparrow$} & \textbf{Cost\,$\downarrow$} & \textbf{Saf.$_{\min}$\,$\uparrow$} & \textbf{Saf.$_{\max}$\,$\uparrow$} \\
\midrule
\rowcolor{green!20}
\textbf{Full model} & \textbf{11.36} & \textbf{2.88} & \textbf{0.96} & \textbf{0.98} \\
No trav.\ input     & 8.37           & 3.42          & 0.93          & 0.98          \\
No $s_g$ input      & 11.26          & 3.37          & 0.95          & 0.98          \\
No cond.\ emb.      & 5.83           & 5.34          & 0.90          & 0.97          \\
\bottomrule
\end{tabular}
\label{tab:path_ablation}
\end{table}

\subsubsection{Diffusion Policy Design:}

We evaluate two critical design decisions: guidance strategy and generative formulation. As shown in Fig.~\ref{fig:design_choices}, our approach substantially outperforms alternatives.
Conditional traversability guidance achieves 11.36 clearance and 2.88 cost, while classifier-free guidance (CFG) degrades to 5.20 clearance and 5.43 cost with 5$\times$ slower inference (0.48 s vs.\ 90\,ms). CFG's double forward pass and signal amplification, designed for weak text conditioning, prove detrimental when strong multi-modal conditioning (traversability, goal, embodiment) is already present. For generative formulations, we compare $\varepsilon$-prediction (noise prediction) against sample prediction (X-pred.), which directly predicts the clean trajectory $x_0$. The former maintains 11.36 clearance, while the latter collapses to 2.95 (74\% degradation). Since our synthetic trajectories are heuristic rather than expert-optimal, $\varepsilon$-prediction learns the underlying distribution and deviates from imperfect supervision, outperforming the direct MSE fitting of sample prediction.

\subsubsection{Denoising Process:}

\begin{figure}[h]
\centering
\includegraphics[width=\columnwidth]{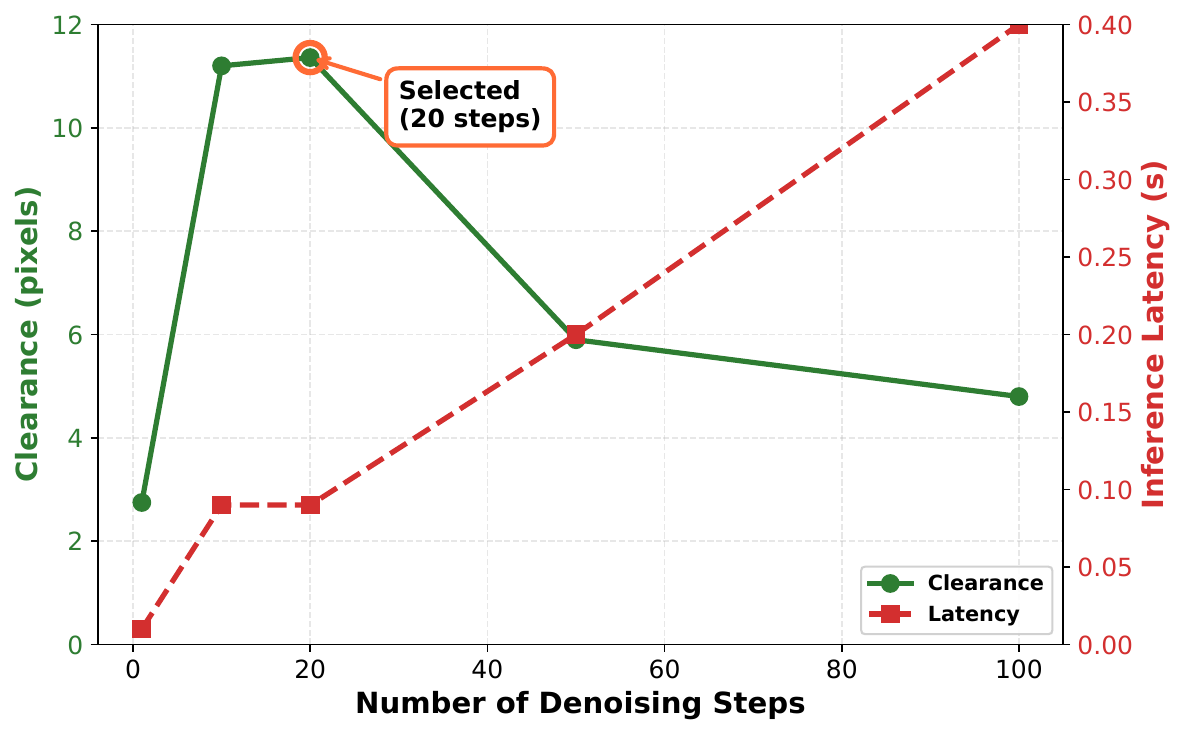}
\caption{Denoising step analysis. We select 20 steps (highlighted) as this setting achieves peak clearance (11.36) with real-time latency (90\,ms).}
\label{fig:denoising_ablation}
\end{figure}

Figure~\ref{fig:denoising_ablation} shows that clearance peaks at 20 steps (11.36 pixels, 90\,ms), balancing quality and speed. Single-step inference yields poor performance (2.75 clearance), while 100 steps provide marginal gains at 4$\times$ latency cost.

\subsection{Baselines}

As no existing method simultaneously performs traversability estimation and diffusion-based trajectory generation across heterogeneous embodiments, we establish two baselines for controlled comparison. First, we adapt DTG~\cite{DTG} by retraining with front-view RGB and manually annotated expert trajectories (11{,}500 aerial, 8{,}000 multi-embodiment), isolating the effect of our planner-free synthetic supervision against dense expert demonstrations. Second, we implement Artificial Potential Fields + VLM, which combines the AnyTraverse teacher model for traversability estimation with classical gradient-based trajectory planning, allowing us to compare learned diffusion against gradient-based optimization on identical traversability inputs.

\subsection{Simulation Deployment}

At deployment, the trajectory heatmap is projected onto the corresponding depth point cloud to extract 3D waypoints, which are then converted to robot base frame coordinates and sent to the controller for trajectory tracking.

We validate the approach in simulation across five diverse scenarios with varying obstacle densities. As shown in Table~\ref{tab:deployment_results}, the model achieves 80\% success rate with 0.33\,m average clearance despite being trained exclusively on real-world RGB data. The performance gap compared to real deployment (80\% vs. 100\% in simpler scenarios) stems from higher noise and artifacts in simulated depth sensors versus real RealSense data, suggesting future work could predict depth directly along candidate trajectories from RGB input.

\subsection{Real-World Deployment}

We evaluate on two platforms. The UAV is a custom 8-inch quadrotor running
ArduPilot with MAVROS high-level control, onboard state estimation via
OpenVINS~\cite{9196524} with a RealSense T265, and off-board trajectory
generation. The ground robot is a Unitree Go1 quadruped with a front-mounted
RealSense D435i, running inference and control onboard on an NVIDIA Jetson
Orin AGX for velocity-based tracking via the robot SDK. Notably, the pipeline
requires neither global localization nor pre-built maps: trajectories are
egocentric and regenerated each inference cycle in the robot's current camera
frame.

\textbf{Trajectory projection and safety.} At each inference cycle, the diffusion model jointly predicts trajectory heatmaps and a traversability map, where the traversability prediction guides the denoising process toward safe and navigable regions. The traversability map is additionally used to estimate the local proportion of traversable area around the robot, helping avoid cornering and dead-end situations.

Near-field trajectory predictions are typically geometrically consistent, while uncertainty increases for longer-range predictions due to ambiguity in monocular RGB depth perception. To prevent this uncertainty from propagating into execution, predicted trajectory heatmaps are projected onto the RealSense depth point cloud to recover 3D waypoints in the robot base frame. Because the diffusion model produces only a small set of trajectory candidates, a localized Artificial Potential Field (APF) cost is evaluated only on this candidate set rather than over the full point cloud. Each candidate is scored using goal progress and obstacle clearance, and the best-scoring trajectory is executed. This lightweight selection mechanism prevents long-range depth ambiguity from propagating into the executed 3D path. The 90\,ms inference cycle further enables continuous replanning, allowing transient errors to be corrected in subsequent planning steps.

\subsubsection{ Comparison with Baselines:}

\begin{figure*}[!t]
\centering
\setlength{\tabcolsep}{1.5pt}
\renewcommand{\arraystretch}{0}

\newcommand{\cellW}{2.6cm}
\newcommand{\cellH}{1.95cm}   
\newcommand{\imgC}[1]{%
  \includegraphics[width=\cellW,height=\cellH]{#1}%
}
\newcommand{\rowLbl}[1]{%
  \raisebox{0.55cm}{\rotatebox{90}{\small\bfseries #1}}%
}

\begin{minipage}[t]{0.49\textwidth}
\centering
\textbf{UAV (Aerial, 50\% data)}\\[0.4mm]
\begin{tabular}{@{}c@{\hspace{1pt}}c@{\hspace{1pt}}c@{\hspace{1pt}}c@{}}
\rowLbl{DTG\,\cite{DTG}} &
\imgC{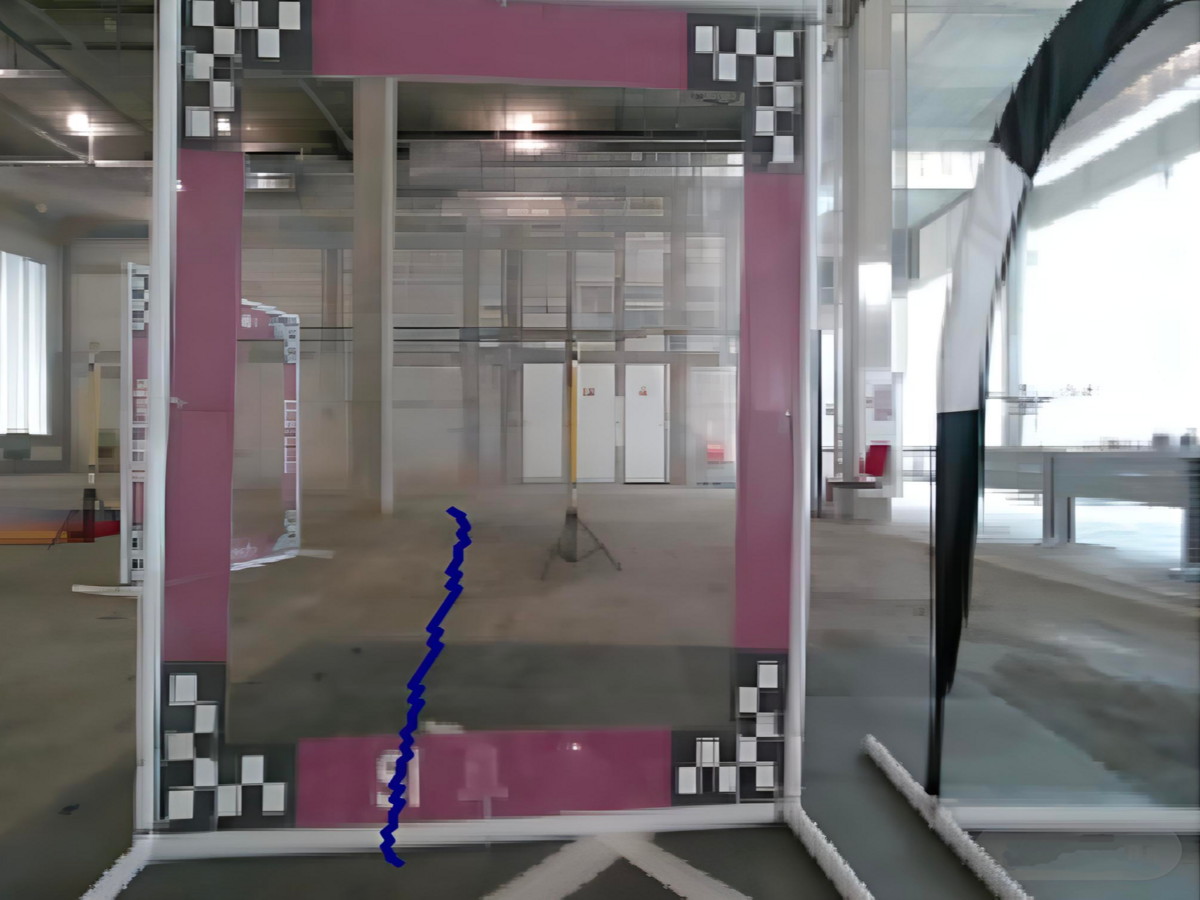} &
\imgC{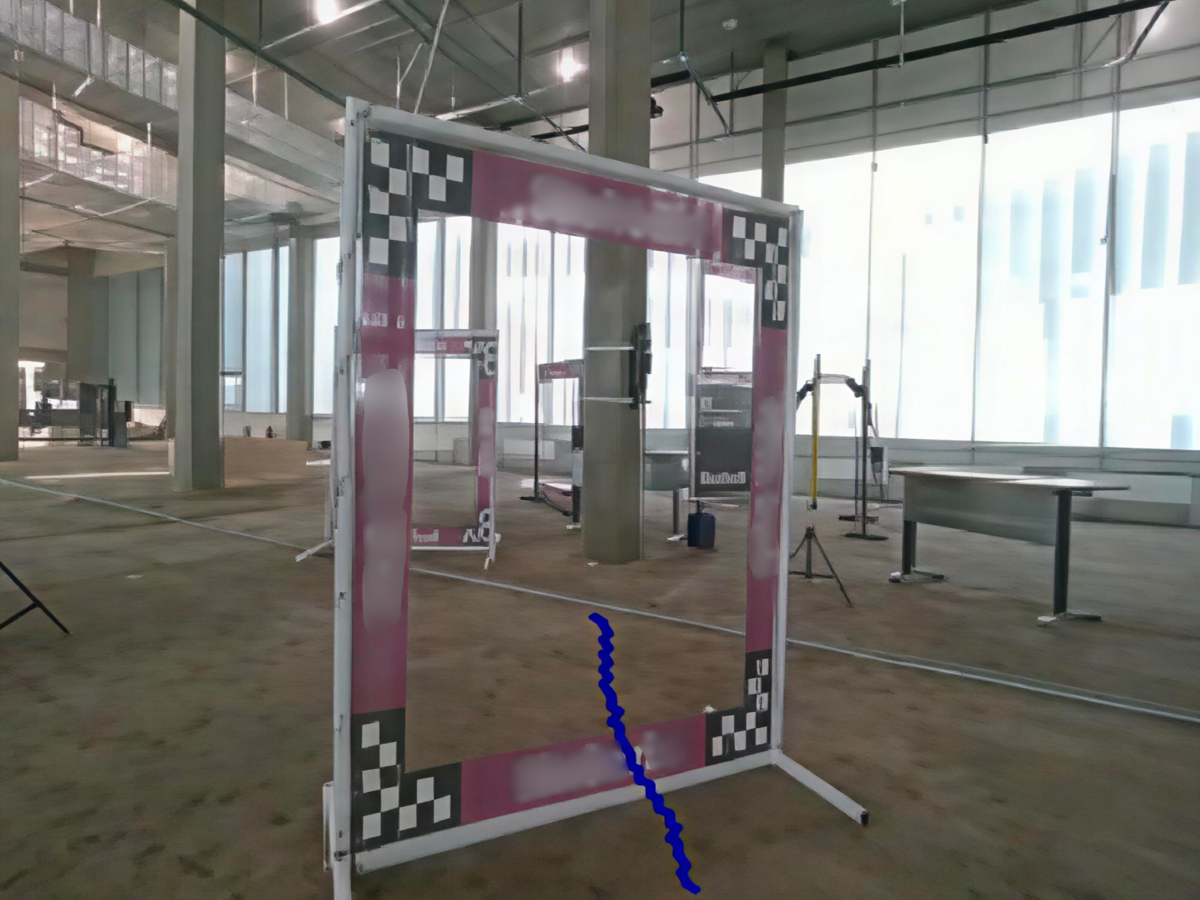} &
\imgC{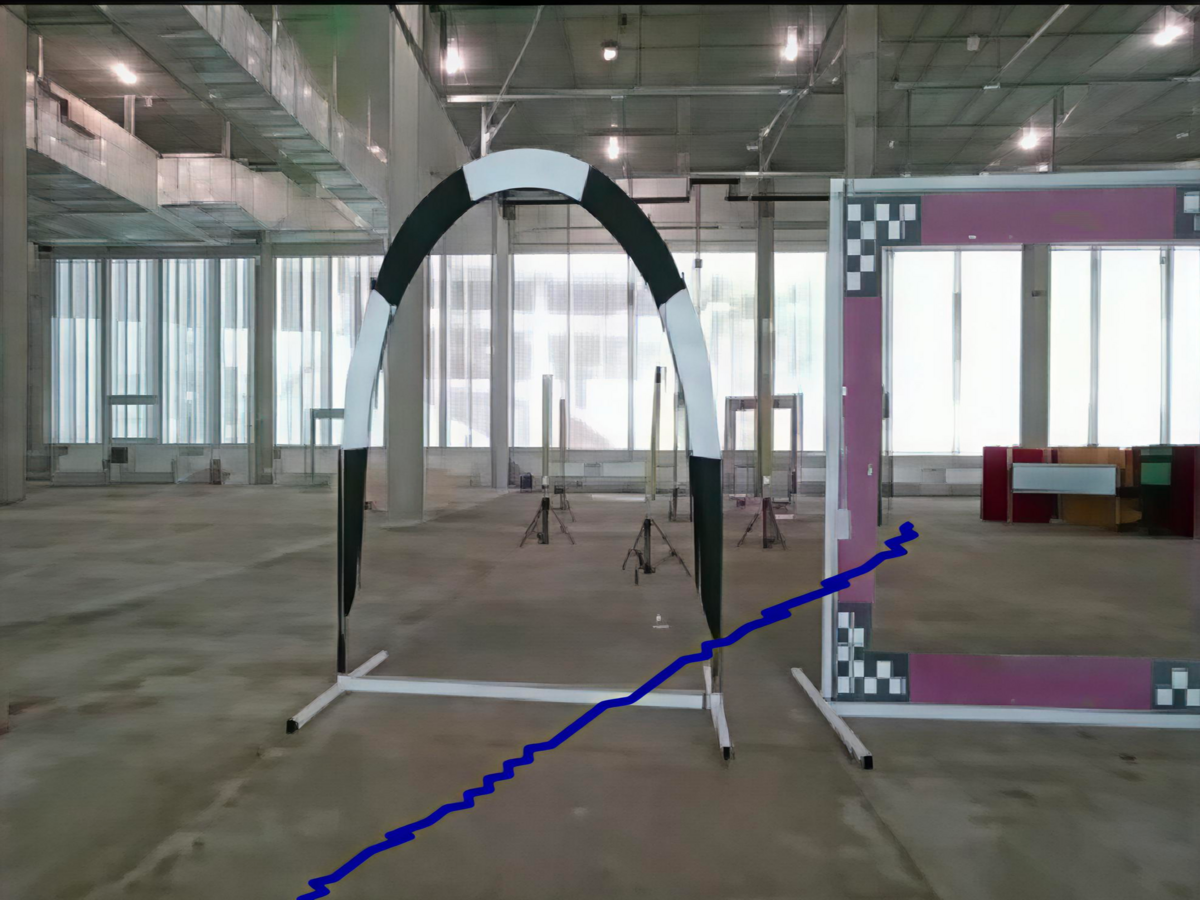} \\
\rowLbl{Ours} &
\imgC{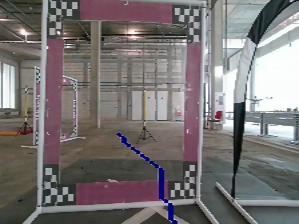} &
\imgC{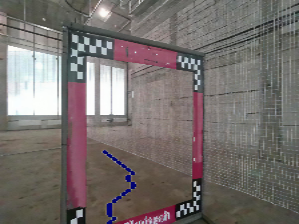} &
\imgC{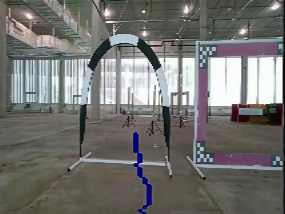} \\
\rowLbl{Trav.\ map} &
\imgC{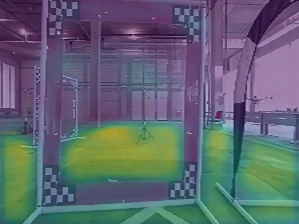} &
\imgC{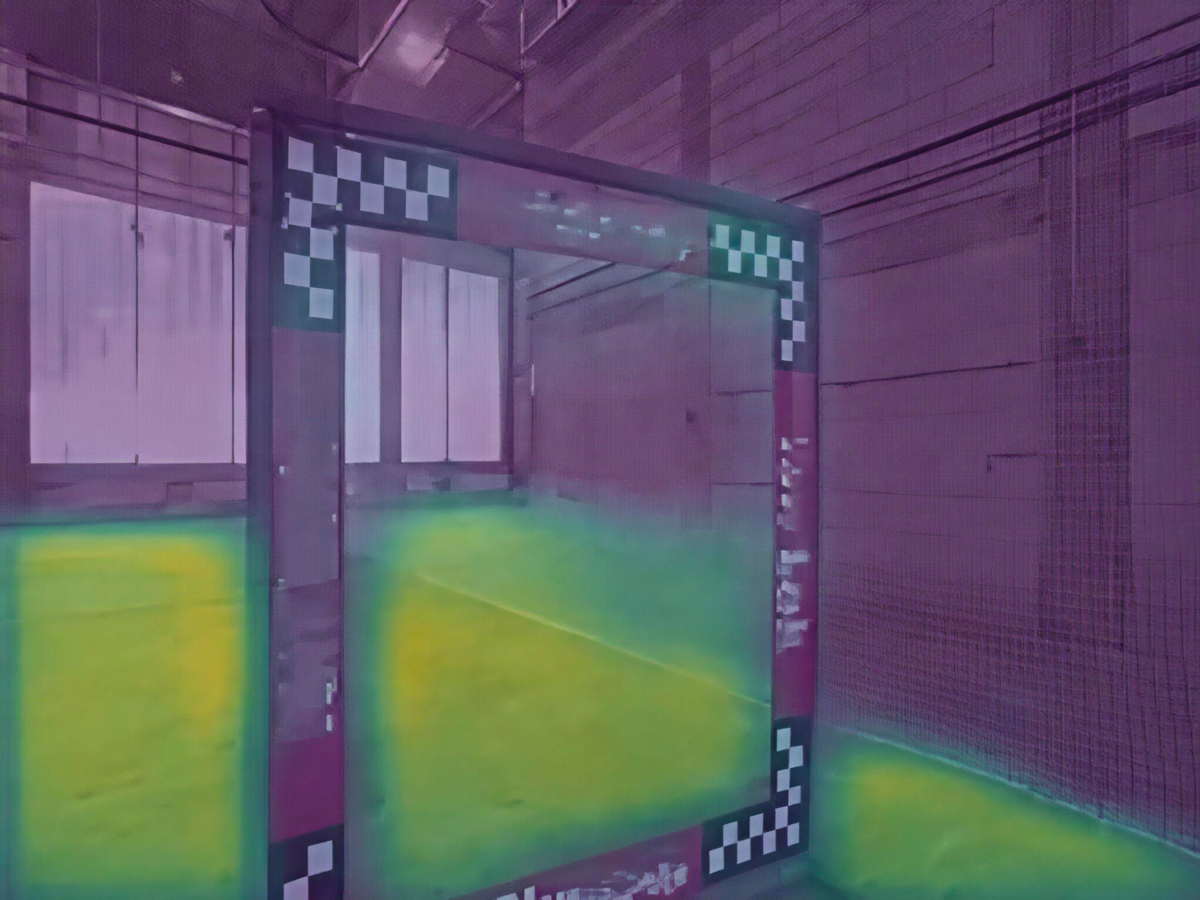} &
\imgC{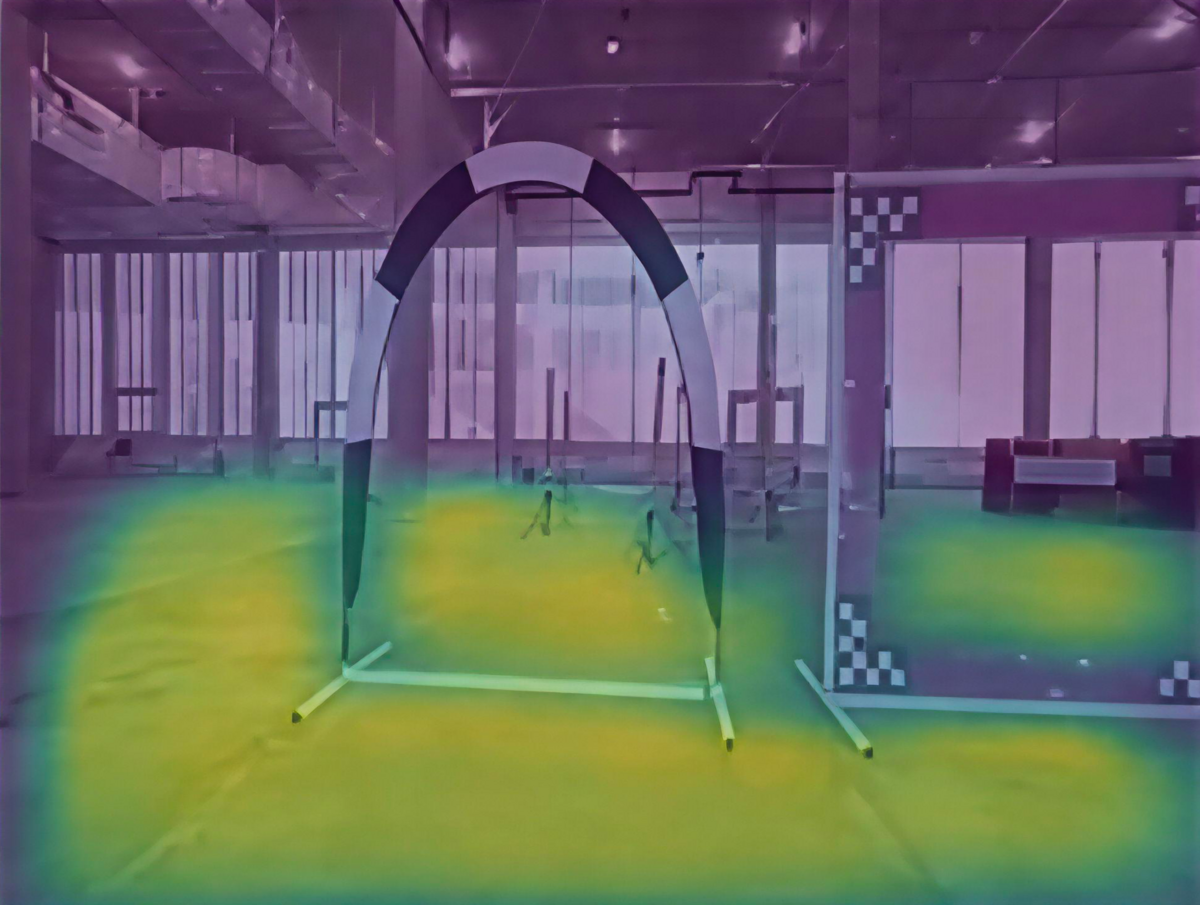} \\
\end{tabular}
\end{minipage}
\hfill
\begin{minipage}[t]{0.49\textwidth}
\centering
\textbf{UGV (Quadruped, 50\% data)}\\[0.4mm]
\begin{tabular}{@{}c@{\hspace{1pt}}c@{\hspace{1pt}}c@{\hspace{1pt}}c@{}}
\rowLbl{DTG\,\cite{DTG}} &
\imgC{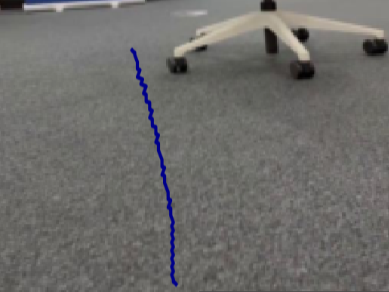} &
\imgC{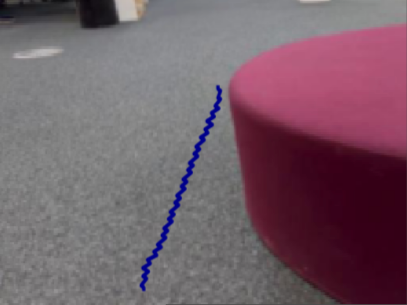} &
\imgC{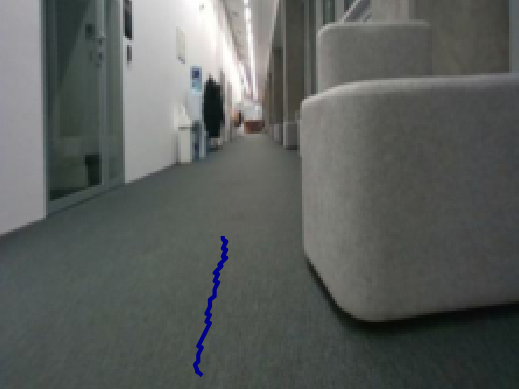} \\
\rowLbl{Ours} &
\imgC{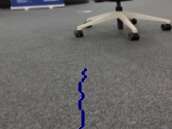} &
\imgC{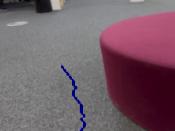} &
\imgC{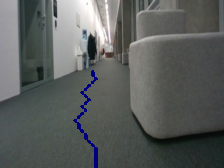} \\
\rowLbl{Trav.\ map} &
\imgC{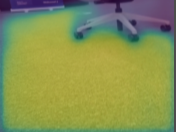} &
\imgC{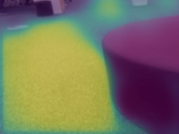} &
\imgC{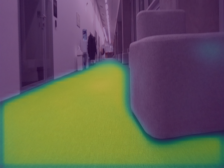} \\
\end{tabular}
\end{minipage}

\caption{Qualitative comparison between DTG~\cite{DTG} baseline and \textit{EmbodiedDiffusion} on aerial and quadruped platforms, with our predicted traversability map shown in the bottom row. A revealing case appears in the UAV columns containing both a black-and-white and a purple gate: DTG consistently selects the same purple gate across scenes because it was trained to imitate expertly labeled ground-truth trajectories that always passed through that specific gate. \textit{EmbodiedDiffusion}, which learns the distribution of feasible paths via traversability regularization rather than memorizing specific trajectories, recognizes any narrow opening as traversable and selects gates based on the predicted traversability map. This directly demonstrates that distribution-level learning generalizes beyond the specific examples seen during training.}
\label{fig:drone_dog_baseline_comparison}
\end{figure*}

Figure~\ref{fig:drone_dog_baseline_comparison} shows our method generates safe trajectories comparable to the supervised baseline without ground-truth annotations, while maintaining better cross-embodiment consistency through traversability-guided diffusion.

\newcommand{\tile}{0.175\textwidth}

\newcommand{\img}[1]{\includegraphics[width=\tile]{#1}}

Table~\ref{tab:baseline_comparison} reveals platform-dependent baseline performance:
Artificial Potential Fields succeeds on UGV (100\%) but fails on UAV
(50\%), while DTG achieves 60--75\% success with expert
supervision. Under this matched-comparison protocol, our method achieves
89\% success on the UAV and 100\% on the UGV at the fastest inference
(90\,ms), outperforming DTG (60--75\%) on identical platforms without
any expert supervision.

A particularly informative qualitative observation supports 
the distribution-learning argument from Section~\ref{sec:synthes}. In the
UAV scenes containing two visually distinct gates (black-and-white and purple), DTG consistently selects the 
purple gate across all evaluations because its training 
trajectories — manually annotated and expertly executed — 
always passed through that specific gate. The model 
effectively memorizes the gate identity rather than 
learning the underlying concept of traversable openings. 
In contrast, \textit{EmbodiedDiffusion} selects gates 
based on the predicted traversability map, recognizing 
either gate as a valid traversable passage. This 
behavioral difference, visible in Fig.~\ref{fig:drone_dog_baseline_comparison}, 
illustrates the practical generalization benefit of 
distribution learning over imitation: our method does not 
need to encounter every specific obstacle configuration 
during training to handle it correctly at deployment.

\begin{table}[t]
\centering
\caption{Baseline Comparison across Heterogeneous Platforms.}
\setlength{\tabcolsep}{3.2pt}
\renewcommand{\arraystretch}{1.05}
\begin{tabular}{lccccc}
\toprule
\textbf{Method} &
\textbf{Time (ms)$\downarrow$} &
\textbf{Clear.$\uparrow$} &
\textbf{Safe$_{\min}\uparrow$} &
\textbf{Safe$_{\max}\uparrow$} &
\textbf{Success$\uparrow$} \\
\midrule
\multicolumn{6}{l}{\textit{UAV (Aerial Robot)}} \\
\midrule
DTG  & 240 & 4.85 & 0.26 & 0.69 & 60\% \\
APF + VLM & 600 & 3.56 & 0.29 & 0.37 & 50\% \\
\textbf{Ours} & \cellcolor{green!20}\textbf{90} & \cellcolor{green!20}\textbf{8.75} & \cellcolor{green!20}\textbf{0.50} & \cellcolor{green!20}\textbf{0.80} & \cellcolor{green!20}\textbf{89\%} \\
\midrule
\multicolumn{6}{l}{\textit{UGV (Quadruped Robot)}} \\
\midrule
DTG & 240 & 8.67 & 0.55 & 0.83 & 75\% \\
APF + VLM & 160 & 10.31 & 0.90 & 0.97 & 100\% \\
\textbf{Ours} & \cellcolor{green!20}\textbf{90} & \cellcolor{green!20}\textbf{10.45} & \cellcolor{green!20}\textbf{0.72} & \cellcolor{green!20}\textbf{0.85} & \cellcolor{green!20}\textbf{100\%} \\
\bottomrule
\end{tabular}
\label{tab:baseline_comparison}
\end{table}

In addition to the baseline comparison, our real-world experiments evaluate the proposed framework across heterogeneous platforms (a quadruped and an aerial robot) along
three deployment-oriented dimensions:
\begin{itemize}
\item \textit{Embodiment conditioning}: verifying that generated
trajectories are physically valid for each platform.
\item \textit{Data efficiency}: assessing performance under reduced
embodiment-specific training data.
\item \textit{Cross-scene generalization}: evaluating transfer to unseen indoor environments.
\end{itemize}

For each deployment scenario, we report success rate and minimum obstacle clearance 
(from RealSense point clouds). Success requires reaching the goal without collision 
or intervention. Each scenario comprises five trials with averaged metrics.

\subsubsection{Embodiment Conditioning:}
\label{sec:embody}
Table~\ref{tab:deployment_results} shows successful 
cross-platform deployment: quadruped achieves 0.41--1.19\,m
clearance (ground-based spatial freedom) while aerial robot
maintains 0.31--0.45\,m in tighter corridors. Both 
achieve 67--100\% success, demonstrating that embodiment 
conditioning captures platform-specific constraints without 
separate models.

The body-specific state vector, combined with visually 
similar training environments shared across platforms, 
allows the model to distinguish between embodiments and 
generate platform-appropriate behaviors. This is most 
evident in the quadruped trajectories: rather than 
attempting to navigate through narrow openings as an 
aerial robot would, the quadruped consistently plans 
routes around tight areas, reflecting its ground-based 
spatial constraints and wider physical footprint. This platform-specific behavior emerges purely from the FiLM-modulated 
state conditioning without any explicit rule encoding or 
separate model per platform.

\subsubsection{Data Efficiency:}

We evaluate the quadruped across three training data regimes
(100\%, 50\%, 30\%). As shown in
Table~\ref{tab:deployment_results}, the model maintains
100\% navigation success under both full and half-data settings,
achieving obstacle clearances of 1.19\,m and 0.41\,m,
respectively. Performance degrades at 30\% training data
(67\% success, 0.56\,m clearance), suggesting an effective
generalization threshold near 50\% of the original dataset.
These results indicate that the distilled VLM priors provide
strong semantic and geometric regularization, enabling robust
traversability prediction even under substantially reduced
supervision. 
\begin{table}[h]
\footnotesize
\setlength{\tabcolsep}{3pt}
\renewcommand{\arraystretch}{1.1}
\centering
\caption{Real-World Deployment Results across Three Settings: UGV Data Efficiency, UAV Cross-Scene Generalization, and UAV Sim-to-Real Transfer.}
\label{tab:deployment_results}
\begin{tabular}{@{}llcc@{}}
\toprule
\textbf{Experiment} & \textbf{Setting} & 
\textbf{Success~$\uparrow$} & \textbf{Clearance~(m)~$\uparrow$} \\
\midrule
\multicolumn{4}{l}{\textit{UGV --- data efficiency 
(quadruped robot)}} \\
\midrule
\rowcolor{green!20}
UGV & 100\% data & \textbf{100\%} & \textbf{1.19} \\
\rowcolor{green!20}
UGV & 50\% data  & \textbf{100\%} & 0.41 \\
UGV & 30\% data  & 67\%           & 0.56 \\
\midrule
\multicolumn{4}{l}{\textit{UAV --- cross-scene 
generalization (aerial robot)}} \\
\midrule
\rowcolor{green!20}
UAV & Single gate (G)         & \textbf{100\%} & 0.45 \\
UAV & Sequential gates (2G) & 67\% & 0.31 \\
\rowcolor{green!20}
UAV & Gate + obstacle (GO)    & \textbf{100\%} & 0.40 \\
\midrule
\multicolumn{4}{l}{\textit{UAV --- simulation-to-real 
transfer}} \\
\midrule
\rowcolor{green!20}
UAV Sim & Five sim scenarios  & 80\%           & 0.33 \\
\bottomrule
\end{tabular}
\end{table}
\subsubsection{Cross-Scene Generalization:}\label{sec:cross-scene}We evaluate the aerial robot on three unseen scenarios: single gate, gate with novel
obstacle, and sequential gates (2G). Across all scenarios, the gates were further 
varied by occluding them with different types of objects 
placed both in front of and behind the gate, testing 
whether the framework can correctly identify the 
traversable opening despite changes in surrounding visual 
context. The drone achieves 100\% success (0.45\,m and 
0.40\,m clearance) on the single-gate and gate-with-obstacle 
scenarios, and drops to 67\% (0.31\,m clearance) on the sequential-gate (2G) scenario due
to tighter spatial constraints.
Robustness to foreground and background occlusion further 
supports the generalization argument from 
Section~\ref{sec:general}: the framework 
recognizes traversability based on the geometry of the 
opening rather than memorizing specific scene layouts.

Failures were defined as clearance below 0.2\,m rather than actual collisions, occurring when conservative human intervention was triggered by depth sensor distortion that degraded close-range distance estimation.

\section{Limitations and Future Work}
The framework generates trajectory heatmaps in image space, which
are projected onto depth point clouds for 3D waypoint extraction.
This step introduces sensitivity to depth sensor noise: in simulation,
degraded point cloud quality reduced success rate to 80\% compared to
100\% in real-world conditions, and robustness remains limited in
environments with reflective surfaces or sensor occlusion.
We identify three directions for future development.

\noindent\textbf{BEV Trajectory Diffusion from Temporal RGB.}
We plan to migrate the trajectory representation from
image-space heatmaps to a Bird's Eye View (BEV) representation
built from temporal RGB sequences and supervised during training
with depth-based BEV maps. Using multi-frame cues to recover spatial
structure removes the dependence on depth projection, enabling
metric-scale planning from monocular RGB alone and improving
robustness to sensor noise and occlusion.

\noindent\textbf{Semantic-Aware Goal Selection.}
Unlike most diffusion-based planners, \textit{EmbodiedDiffusion}
requires no continuous intermediate goal input, instead selecting
the most traversable region as the target. We will extend this with
DINOv2 attention maps to autonomously identify semantically meaningful
targets such as doorways or docking stations, further reducing
supervision without explicit object detection.

\noindent\textbf{Online Traversability Refinement.}
The distilled VLM priors provide strong semantic traversability estimates, letting the robot start from rich prior world knowledge rather than sparse online feedback and incrementally refine its understanding through onboard observation and physical interaction. Future work will explore refinement using onboard sensing such as IMU and proprioceptive feedback, enabling adaptation to finer body-dynamic constraints and surface properties beyond what RGB alone can provide.

\section{Conclusion}

We presented \textit{EmbodiedDiffusion}, a unified traversability-guided diffusion framework that simultaneously predicts traversability and generates feasible trajectories directly from RGB observations, enabling embodiment-conditioned navigation without expert demonstrations, explicit geometric maps, or deployment-time prompt engineering. By distilling VLM-based traversability reasoning into a compact diffusion model, the proposed approach learns navigation as a distribution-level traversability problem rather than memorization of expert trajectories.

Experimental results across quadruped and aerial platforms demonstrate that planner-free synthetic supervision is sufficient to achieve robust real-world navigation performance, while lightweight FiLM-based embodiment conditioning enables rapid cross-platform adaptation from only 10\,min of data. The proposed end-to-end formulation further improves UAV obstacle clearance by up to 80\% over the strongest baseline while operating at 6.7$\times$ faster inference.

Together, these results show that a single lightweight onboard model can unify traversability reasoning and trajectory generation across heterogeneous robotic embodiments, providing a scalable foundation for real-time navigation in complex and previously unseen environments.


\bibliographystyle{IEEEtran}
\bibliography{references}

\end{document}